%% file: main.tex
\documentclass[a4paper]{styles/svproc}
\usepackage{url}

\usepackage[bookmarks=true]{hyperref} 
\usepackage{import}
\usepackage{amsmath, amssymb}
\usepackage{graphicx} 
\usepackage{subfigure}
\usepackage{multirow}
\usepackage{multicol}
\usepackage[symbol]{footmisc}

\begin{document}
\mainmatter              
\title{Contact-Implicit Planning and Control for Non-Prehensile Manipulation Using State-Triggered Constraints}
\titlerunning{Contact-Implicit Planning and Control of Pushing Using STCs}  

%
\author{Maozhen Wang\footnote[1]{Indicates equal contribution}\inst{1}\footnote[2]{Currently at Amazon Robotics, MA, USA} \and Aykut \"{O}zg\"{u}n \"{O}nol$^{*}$\inst{1}\footnote[3]{Currently at Toyota Research Institute, Cambridge, MA, USA}
Philip Long\inst{2}\and Ta\c{s}k\i n Pad\i r\inst{1}
\thanks{This material is based upon work supported by the National Science Foundation under Award No. 1928654.
The authors would like to thank Dr. Michael Szmuk, Dr. Taylor P. Reynolds, and Prof. Beh\c{c}et A\c{c}{\i}kme{\c{s}}e of the Autonomous Controls Lab at the University of Washington for their useful feedback and insights.}
}
\authorrunning{M. Wang, A. \"{O}nol, P. Long, \& T. Pad\i r} 
%
\tocauthor{Maozhen Wang  Aykut \"{O}zg\"{u}n \"{O}nol
Philip Long and Ta\c{s}k\i n Pad\i r}
\institute{Institute for Experiential Robotics, Northeastern University, Boston, MA, USA\\
\and
Atlantic Technological University, Galway, Ireland \\
}

\maketitle              

\import{sections/}{0_abstract.tex}
\import{sections/}{1_intro.tex}
\import{sections/}{2_background.tex}
\import{sections/}{3_approach.tex}

\import{sections/}{4_results.tex}

\import{sections/}{5_conclusion.tex}

\bibliographystyle{styles/bibtex/spmpsci}
\bibliography{refs}

\end{document}

%% file: sections/0_abstract.tex
\begin{abstract}
We present a contact-implicit planning approach that can generate contact-interaction trajectories for non-prehensile manipulation problems without tuning or a tailored initial guess and with high success rates. This is achieved by leveraging the concept of state-triggered constraints (STCs) to capture the hybrid dynamics induced by discrete contact modes without explicitly reasoning about the combinatorics. STCs enable triggering arbitrary constraints by a strict inequality condition in a continuous way. We first use STCs to develop an automatic contact constraint activation method to minimize the effective constraint space based on the utility of contact candidates for a given task. Then, we introduce a re-formulation of the Coulomb friction model based on STCs that is more efficient for the discovery of tangential forces than the well-studied complementarity constraints-based approach. Last, we include the proposed friction model in the planning and control of quasi-static planar pushing. The performance of the STC-based contact activation and friction methods is evaluated by extensive simulation experiments in a dynamic pushing scenario. The results demonstrate that our methods outperform the baselines based on complementarity constraints with a significant decrease in the planning time and a higher success rate. We then compare the proposed quasi-static pushing controller against a mixed-integer programming-based approach in simulation and find that our method is computationally more efficient and provides a better tracking accuracy, with the added benefit of not requiring an initial control trajectory. Finally, we present hardware experiments demonstrating the usability of our framework in executing complex trajectories in real-time even with a low-accuracy tracking system.

\keywords{contact modeling, manipulation planning, optimization}
\end{abstract}

%% file: sections/1_intro.tex
\section{Introduction}
Non-prehensile manipulation will be a key capability for robots in both home and industrial environments~\cite{stuber2020let}. However, there is still a lack of reliable methods that can compose contact-rich motions given only a high-level goal. The main challenge is that the planning and control of contact interactions require discrete decisions concerning the time and location of contact mode transitions that impose switching constraints altering the evolution of the system dynamics. To avoid predefined contact schedules, a hybrid problem using mixed-integer programming (MIP) may be solved, however, explicit modeling can be prohibitive as the complexity grows exponentially with discrete variables~\cite{posa2014direct}.
Fortunately, many discrete elements can be expressed in terms of continuous variables -- such as the distance, velocity, and force in the contact frame -- by using complementarity constraints (CCs) which provide an efficient way of including bi-directional conditions in programs without the combinatorial complexity of explicitly doing so. Thus, contact dynamics is defined as a smooth optimization problem with CCs in many physics engines and planning algorithms~\cite{stewart1996implicit,yunt2005trajectory,posa2014direct}. Yet, describing the rich discrete aspects of contact-related problems using only bi-directional statements is restrictive and may be inefficient. We hypothesize that a continuous model that implies uni-directional \textit{if} conditions would be useful as it would provide a more modular building block. Hence, we propose the use of state-triggered constraints (STCs), first put forward by Szmuk et al.~\cite{szmuk2020successive}, and analyze their performance for contact-implicit planning (CIP) through non-prehensile manipulation examples.

\subsection{Contributions}
The main contributions of our work are:
\begin{itemize}
    \item An automatic contact constraint activation (CA) method that triggers \textit{only} the contact constraints useful for given task, thus reducing the problem's sensitivity to the number of contact pairs and enabling automatic contact candidate assignment, i.e., all perceived surfaces can be assigned as contact candidates as they are handled efficiently in the optimization. Extensive simulations show that the proposed CA method improves the computational efficiency and the success rate.
    \item We reformulate the Coulomb friction model using STCs and compare it to a baseline based on CCs for discovering tangential forces to rotate a box to random goal orientations. The proposed model is significantly faster and more robust than the baseline.
    \item The STC-based friction model is used to model the contact modes of planar quasi-static pushing and compared to a MIP-based controller. Our method outperforms the baseline both in speed and accuracy. Moreover, the method can plan contact-interaction trajectories given only a desired state trajectory without any heuristics. Although our method may run slower than the learning-based variant~\cite{hogan2020reactive}, we demonstrate experimentally that our STC-based control can plan and track complex trajectories at a frequency high enough for real-time applications.
\end{itemize}
To the best of our knowledge, this is the first usage of STCs in contact modeling. We show that STCs hold immense promise for modeling discrete elements in this domain and can mitigate the need for explicit combinatorics and heuristics. Our proposed CA and friction methods serve as example applications for STCs. The extensive simulation and hardware experiments prove the efficacy of the proposed methodology.


\subsection{Related Work}
\subsubsection{Mixed-Integer Programming}
The planning and control of contact-rich motions can be achieved by building a hybrid problem with explicit discrete variables and then solved by a MIP approach. In~\cite{deits2014footstep,aceituno2017simultaneous}, MIP-based planners are developed and applied to locomotion on uneven terrains. In manipulation, MIP~\cite{hogan2016feedback,aceituno2020global} and exhausted tree search~\cite{doshi2020hybrid} have been used to achieve desired object behaviors. Nevertheless, these approaches typically require simplifications and/or heuristics to make them computationally tractable. \cite{hogan2020reactive,marcucci2020warm,driess2020deep} introduced various initialization strategies to reduce the computational burden, mostly with data-driven techniques. 

\subsubsection{Contact-Implicit Trajectory Optimization}
Alternatively, contact-interaction trajectories can be planned by incorporating a differentiable contact model into a trajectory optimization framework, i.e., contact-implicit trajectory optimization (CITO). \cite{todorov2012discovery,todorov2012manipulation} have shown complex contact-rich behaviors can be synthesized for animated characters using convex optimization with soft constraints by sacrificing physical realism. In robotics applications, where motions must be physically accurate, CCs are widely used to model inelastic rigid-body contacts with Coulomb friction. In~\cite{yunt2005trajectory} the non-smooth trajectory optimization problem with impacts and discontinuities are transcribed into a bi-level program with CCs. Posa et al.~\cite{posa2014direct} solved this problem simultaneously using direct collocation. Similar methods have been proposed for planar manipulation~\cite{gabiccini2018computational} and dynamic pushing~\cite{sleiman2019contact}. In \cite{mastalli2016hierarchical,marcucci2017two}, hierarchical strategies using warm-starting to improve the computational efficiency are presented, while \cite{manchester2020variational,patel2019contact} propose methods to improve the integration accuracy of such numerical schemes. Other works have focused on replacing CCs for more efficient direct programs, e.g., \cite{buchli2018gait,chatzinikolaidis2020contact,stouraitis2020multi}.

The optimization problem can be solved faster than direct optimization via Hessian-approximating differential dynamic programming (DDP) variants, such as iterative linear quadratic regulator (iLQR)~\cite{todorov2004ilqr}, typically with smooth contact models as these approaches cannot easily handle constraints. In \cite{todorov2012synthesis,buchli2016relax}, smoother fragments of CCs are used with DDP variants, and the potential to run CITO as model predictive control (MPC) is demonstrated. \cite{buchli2018npmc} utilized an explicit smooth contact model to enable real-time MPC for highly-dynamic quadruped motions. However, these methods usually require either a good initial guess and/or tedious tuning due to the unreliable convergence of DDP methods. Recently, we proposed a CITO framework based on a variable smooth contact model~\cite{onol2018comparative} that decouples the relaxation from the contact model by injecting smooth virtual forces into the underactuated dynamics with frictional contacts. Hence, both nonlinear programming and DDP variants can be used to solve the problem, as shown by \cite{acikmese2016scvx,onol2019contact,wang2020affordance,onol2020tuning}. While the methods developed here are applicable to our previous work, we build upon a CC-based CITO framework similar to \cite{posa2014direct} as it is more widely used in the related literature and thus a more relevant comparison.

\subsubsection{Planar Pushing}
The mechanics of planar pushing in the presence of friction was first studied in \cite{mason1986mechanics} which focuses on the rotational direction and the object's center. In \cite{lynch1996stable}, the mechanics for stable pushing is studied. More recently, \cite{zhou2019pushing} have shown that with a sticking contact, Dubins path can plan trajectories efficiently. In~\cite{hogan2016feedback} a hybrid MPC framework is proposed to track a pushing trajectory without limiting the contact mode to sticking only, instead the contact mode for each time step is defined as a discrete variable. To avoid the combinatorial problem's computational, convex problems are solved for a fraction of potential contact schedules for the prediction horizon, and the solution is determined comparatively. In \cite{hogan2020reactive}, an aggregated contact mode selection method is proposed to reduce the complexity of MIP and to eliminate the need for a predefined contact mode schedule. They proposed a learning method to predict the proper contact mode to achieve a higher control frequency. However, this approach is not very versatile as it requires re-training for object shape variations.

\subsubsection{State-Triggered Constraints}
STCs were first introduced in \cite{szmuk2020successive} as a more modular, uni-directional alternative to CCs and were used to avoid MIP and enable real-time performance for a rocket landing task with state-dependent constraints. In \cite{szmuk2019successive}, this concept is extended into a more general form called compound STCs showing that more articulate trigger and constraint conditions can be obtained by applying Boolean logic operations. STCs have been successfully applied to real-time rocket landing~\cite{reynolds2019state,reynolds2020dual} and quadrotor path planning~\cite{szmuk2019real}.



%% file: sections/2_background.tex
\section{Background}

\subsection{Contact Model}
\label{sec:contact_model}
Following \cite{stewart1996implicit}, a contact impulse due to a frictionless inelastic collision between two rigid bodies can be modeled by $0 \leq \phi(\mathbf{q}) \perp \lambda_n \geq 0$; where the non-negativity condition for the signed distance $\phi(\mathbf{q})$ prevents interpenetration between bodies, and the non-negativity condition for the normal contact force $\lambda_n$ ensures that the contact bodies can only push each other. The frictional forces $\boldsymbol{\lambda}_t \in \mathbb{R}^2$ can be modeled using the relative tangential (slip) velocity in the contact frame $\psi(\mathbf{q}, \dot{\mathbf{q}})$, the friction coefficient $\mu$ and by splitting tangential force into directional components $\lambda_t^+,\lambda_t^- \geq 0$:
\begin{subequations}
    \label{eq:tangent_cc}
    \begin{align}
        & \mathbf{0} \leq (\mu \lambda_n - \lambda_t^+ - \lambda_t^-) \perp \gamma \geq \mathbf{0}, \\
        & \mathbf{0} \leq (\gamma +  \psi(\mathbf{q}, \dot{\mathbf{q}})) \perp \lambda_t^+ \geq \mathbf{0}, \\
        & \mathbf{0} \leq (\gamma -  \psi(\mathbf{q}, \dot{\mathbf{q}})) \perp \lambda_t^- \geq \mathbf{0},
    \end{align}
\end{subequations}
where $\gamma$ is an auxiliary variable corresponding to the absolute value of the slip velocity. It is noteworthy that this formulation is based on the maximization of the dissipation of kinetic energy subject to pyramidical friction cone constraints, see \cite{manchester2020variational} for a derivation. However, the stationarity condition is neglected, and this leads to not being able to counteract external forces to prevent slippage when the slip velocity is zero. Nevertheless, these constraints are sufficient to ensure that the tangential forces selected by the planner are bounded by the friction polyhedron when sticking and counteracting at the boundary when slipping.

\subsection{Contact-Implicit Trajectory Optimization}
The CIP problem can be transcribed into a nonlinear program (NLP) for $N$ time steps of size $h$ using a backwards Euler discretization of the underactuated dynamics for an $n_v \triangleq n_a+n_u$ degrees-of-freedom (DOF) system with $n_a$ actuated and $n_u$ unactuated DOF:
\begin{subequations}
    \label{eq:optimization_problem}
    \begin{align}
        \underset{\mathbf{X},\mathbf{U},\mathbf{F},\mathbf{S}}{\text{minimize }} \sum_{i=1}^{N} w_1 \left \| \dot{\mathbf{q}}_i \right \|_{2}^2 + w_2 \left \| \mathbf{s}_i \right \|_{1}
    \end{align}
subject to:
    \begin{align}
        & \mathbf{q}_{i+1} = \mathbf{q}_i + h \dot{\mathbf{q}}_{i+1}, \ M_{i+1}(\dot{\mathbf{q}}_{i+1}-\dot{\mathbf{q}}_{i}) = h(\mathbf{S}_a \boldsymbol{\tau}_{i} + \mathbf{J}^T_{c,i+1} \boldsymbol{\lambda}_{c,i}-c_{i+1}) \ \forall i, \label{eq:dynamic_transcription_vel} \\
        & \boldsymbol{\tau}_{L} \leq \mathbf{\boldsymbol{\tau}}_i \leq \boldsymbol{\tau}_{U}, \ \mathbf{x}_{L} \leq \mathbf{x}_i \leq \mathbf{x}_{U} \ \forall i, \ \mathbf{x}_{1} = \mathbf{x}_{init}, \ \mathbf{x}_{N+1} = \mathbf{x}_{goal}, \\
        & \boldsymbol{\phi}(\mathbf{q}_{i}), \boldsymbol{\lambda}_{n,i}, \mathbf{s}_i \geq 0 \ \forall i, \\ 
        & \boldsymbol{\phi}(\mathbf{q}_i) \cdot \boldsymbol{\lambda}_{n,i} \leq \mathbf{s}_i \ \forall i, \label{eq:normal_relaxed_cc}
    \end{align}
\end{subequations}
where $\mathbf{q} \in\mathbb{R}^{n_q}$ and $\dot{\mathbf{q}} \in \mathbb{R}^{n_v}$ are the generalized positions and velocities of the system; $\mathbf{M}(\mathbf{q}) \in \mathbb{R}^{n_v \times n_v}$ and $\mathbf{c}(\mathbf{q},\dot{\mathbf{q}}) \in \mathbb{R}^{n_v}$ denote the mass matrix and the Coriolis, centrifugal, and gravitational terms; $\boldsymbol{\tau} \in \mathbb{R}^{n_a}$ is the vector of generalized joint forces; $\mathbf{S}_a \in \mathbb{R}^{n_v \times n_a}$ is the selection matrix for the actuated DOF; and $\boldsymbol{\lambda}_{c} \in \mathbb{R}^{3n_c}$ is the vector of contact forces for $n_c$ contacts that are projected onto the joint space through the contact Jacobian $\mathbf{J}_{c}(\mathbf{q}) : \mathbb{R}^{n_q} \mapsto \mathbb{R}^{3n_c \times n_v}$. Representing the state by $\mathbf{x} \triangleq [\mathbf{q}^T \ \mathbf{\dot{q}}^T]^T \in \mathbb{R}^{n}$, $\mathbf{X} \triangleq [\mathbf{x}_1,...,\mathbf{x}_{N+1}]$, $\mathbf{U} \triangleq [\boldsymbol{\tau}_1,...,\boldsymbol{\tau}_N]$, $\mathbf{F} \triangleq [\boldsymbol{\lambda}_1,..., \boldsymbol{\lambda}_N]$, $\mathbf{S} \triangleq [\mathbf{s}_1,...,\mathbf{s}_{N+1}]$ are the state, control, force, and slack trajectories; and $\boldsymbol{\tau}_L$, $\boldsymbol{\tau}_U$, $\mathbf{x}_L$, and $\mathbf{x}_U$ are the lower and upper control and state bounds. The slack variables $\mathbf{s}$ are introduced to relax the CCs (including those for frictional forces) and penalized in the cost by an exact penalty function, as described in \cite{manchester2020variational,anitescu2005using}. The constraints are evaluated element-wise~\cite{posa2014direct}.

\subsection{Quasi-Static Planar Pushing}
\label{sec:quasi_static_pushing}
In this work, we consider a single point pusher-slider system under the assumptions of quasi-static motion, Coulomb friction, and uniform pressure distribution over the support plane by following the formulation in \cite{hogan2016feedback}. We model the dynamics of planar pushing using an ellipsoidal approximation of the limit surface~\cite{goyal1991planar}. We consider three contact modes between the pusher and the slider: sticking, sliding left, and sliding right. Each mode imposes a different set of friction constraints on the system. The planning problem can then be formulated as an MIP by linearizing the dynamics given desired state and control trajectories and a contact mode schedule. This framework is used to control the system in a receding horizon fashion such that the deviations from the nominal trajectories are minimized. We refer the reader to \cite{hogan2020reactive,hogan2016feedback} for details.



%% file: sections/3_approach.tex
\section{Approach}
\subsection{State-Triggered Constraints}
An STC consists of a strict inequality trigger condition $g(\mathbf{x},\mathbf{u}) < 0$ and an inequality constraint condition $c(\mathbf{x},\mathbf{u}) \leq 0$ such that
\begin{equation}
    \label{eq:stc}
    g(\mathbf{x},\mathbf{u}) < 0 \Rightarrow c(\mathbf{x},\mathbf{u}) \leq 0.
\end{equation}
That is, the constraint $c(\mathbf{x},\mathbf{u}) \leq 0$ becomes active if the trigger condition $g(\mathbf{x},\mathbf{u}) < 0$ is met. To include this logical implication in a smooth optimization problem, it must be represented using continuous variables, by introducing an auxiliary variable $\sigma \in \mathbb{R}_{++}$ and the following constraints
\begin{subequations}
    \label{eq:cstc_intermediate}
    \begin{align}
        0 \leq \sigma \perp (g(\mathbf{x},\mathbf{u}) + \sigma) \geq 0, \label{eq:stc_lcp} \\
        \sigma c(\mathbf{x},\mathbf{u}) \leq 0.
    \end{align}
\end{subequations}
Given $(\mathbf{x},\mathbf{u})$, \eqref{eq:stc_lcp} defines a linear complementarity problem in $\sigma$ which has a unique solution that can be obtained analytically as $\sigma^*=-\text{min}(g(\mathbf{x},\mathbf{u}),0)$. Plugging $\sigma^*$ into \eqref{eq:cstc_intermediate}, we can obtain an STC that is continuous in $(\mathbf{x},\mathbf{u})$ -- i.e., a continuous STC (cSTC) -- as $-\text{min}(g(\mathbf{x},\mathbf{u}),0) \cdot c(\mathbf{x},\mathbf{u}) \leq 0.$, see \cite{szmuk2020successive} for details.

\subsection{Automatic Contact Constraint Activation}

\begin{figure}[h]
  \centering
  \includegraphics[width=0.5\textwidth]{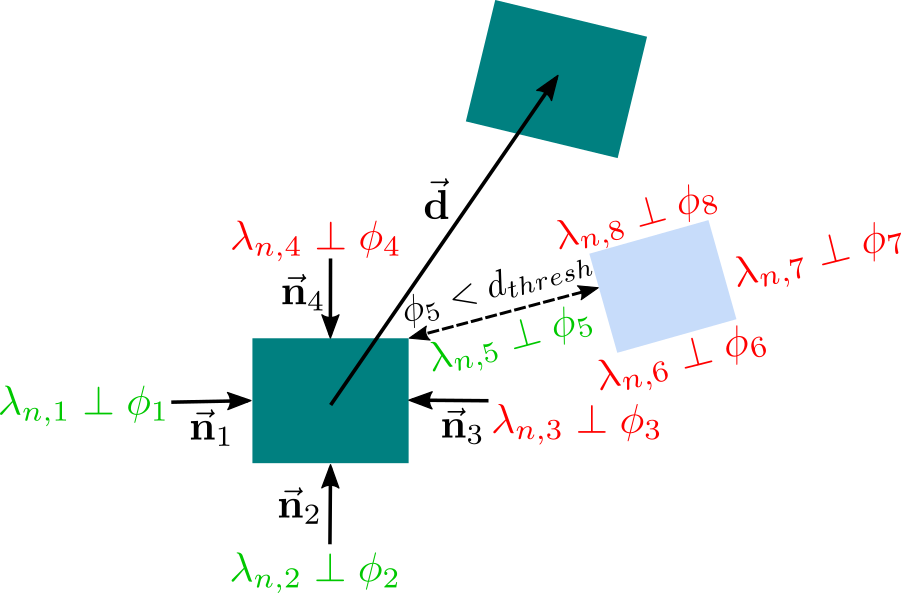}
  \caption{Automatic contact constraint activation: The activated and deactivated contact pairs are indicated by green and red complementarity conditions, respectively. Contact pairs become active if the distance is below a threshold, e.g., the fifth pair in this case.}
  \label{fig:activation}
\end{figure}

In the CC-based CITO, the constraint space grows with the number of contact pairs. Yet, at any instance, only a fraction of contact pairs are useful. Thus, we propose an automatic CA mechanism based on STCs that eliminates unnecessary contact pairs. This concept, based on a simple heuristic for planar pushing, is intended to showcase the versatility of STCs in this domain.

Fig.~\ref{fig:activation} illustrates the CA mechanism for a planar pushing scenario where $\mathbf{d}$ is the vector from the box's current location to the desired. In this case, contact candidates are typically defined by pairing each end effector with each surface of the manipuland. However, the contact pairs that cannot generate normal forces to aid task completion can be deactivated by bounding the corresponding normal forces to zero. This is achieved by replacing the CCs in \eqref{eq:normal_relaxed_cc} by the following STCs
\begin{subequations}
    \label{eq:stc_activation_manipulation}
    \begin{align}
    & g_1 \triangleq \mathbf{N}^T \mathbf{d} > \mathbf{0} \Rightarrow \boldsymbol{\phi}(\mathbf{q}) \cdot \boldsymbol{\lambda}_{n} \leq \mathbf{s}, \\
    & g_2 \triangleq \mathbf{N}^T \mathbf{d} < \mathbf{0} \Rightarrow \boldsymbol{\lambda}_{n} = \mathbf{0},
    \end{align}
\end{subequations}
where $\mathbf{N} \triangleq [\mathbf{n}_1,...,\mathbf{n}_{n_c}] \in \mathbb{R}^{3 \times n_c}$ and $\mathbf{n}$ denotes the surface normal. Note that when the normal force is set to zero, the tangential forces are also suppressed.

In addition to selectively activating contact pairs for manipulation, this mechanism can be used to activate contact constraints efficiently in a cluttered environment. For example, the distance between the manipuland and the environment being less than a threshold, $d_{thresh}$, can be used as the trigger condition to activate important contact pairs on the fly, i.e.,
\begin{subequations}
    \label{eq:stc_activation_environment}
    \begin{align}
    & g_3 \triangleq \boldsymbol{\phi}(\mathbf{q}) < \mathbf{d}_{thresh} \Rightarrow \boldsymbol{\phi}(\mathbf{q}) \cdot \boldsymbol{\lambda}_{n} \leq \mathbf{s}, \\
    & g_4 \triangleq \boldsymbol{\phi}(\mathbf{q}) > \mathbf{d}_{thresh} \Rightarrow \boldsymbol{\lambda}_{n} = \mathbf{0},
    \end{align}
\end{subequations}

This is somewhat similar to the collision constraint relaxation used in \cite{schulman2014motion} when the distance is above a safety margin. It should be noted that the activation of constraints based on task utility, i.e., the addition of \eqref{eq:stc_activation_manipulation} into \eqref{eq:optimization_problem}, is performed only for the contact pairs between the robot and the manipuland. Whereas, the activation of contact pairs that have a distance less than the threshold, i.e., the addition of \eqref{eq:stc_activation_environment} into \eqref{eq:optimization_problem}, is performed for all contact pairs. Thus, contact interactions are enabled between all contact pairs throughout the optimization but in a computationally-efficient way.\footnote{Experiments demonstrating the distance-triggered contact pairs can be found in the accompanying video.}

\subsection{A Reformulation of Coulomb Friction}
\label{sec:stc_friction}
 The Coulomb friction model can be expressed in terms of \textit{if-then} conditions as
\begin{subequations}
    \label{eq:coulomb}
    \begin{align}
        & \psi (\mathbf{q}, \dot{\mathbf{q}}) = 0 \Rightarrow || \boldsymbol{\lambda}_{t} || \leq \mu \lambda_{n}, \\
        & \psi (\mathbf{q}, \dot{\mathbf{q}}) \neq  0 \Rightarrow \lambda_{t} = -sign(\psi (\mathbf{q}, \dot{\mathbf{q}})) \mu \lambda_{n}.
    \end{align}
\end{subequations}
Such logical implications can be conveniently expressed using the following STCs
\begin{subequations}
    \label{eq:tangent_stc}
    \begin{align}
        & g_5 \triangleq \psi (\mathbf{q}, \dot{\mathbf{q}}) > 0 \Rightarrow \lambda_t^- = \mu \lambda_n \text{ and } \lambda_t^+ =0,  \\
        & g_6 \triangleq \psi (\mathbf{q}, \dot{\mathbf{q}}) < 0 \Rightarrow \lambda_t^- =0 \text{ and } \lambda_t^+ = \mu \lambda_n, 
    \end{align}
along with a pyramidical approximation of the boundary constraint
    \begin{align}
    g_7 \triangleq ||\boldsymbol{\lambda}_t||_1 \leq \mu \lambda_n. \label{eq:friction_pyramid}
    \end{align}
\end{subequations}
In contrast to the CC-based formulation in \eqref{eq:tangent_cc},  taking the absolute value of the slip velocity $\gamma$ is not required here, since non-negativity constraints do not need to be enforced when using STCs. Moreover, the Coulomb friction law described in \eqref{eq:coulomb} is expressed sufficiently using only uni-directional implications.
In the remainder, \eqref{eq:tangent_cc} is used for the CC-based friction model, and \eqref{eq:tangent_stc} is used for the STC-based friction model. Note that we use these friction models to model the friction only in contacts between the robot and the manipuland and not for support contacts due to the lack of physical fidelity described in \ref{sec:contact_model}.

\subsection{Quasi-Static Pushing Planning and Execution}
Both of the presented friction models can be employed to capture the switching nature of the quasi-static pushing without discrete modeling such that the MIP-based pushing controller~\cite{hogan2016feedback} can be re-formulated as a direct NLP given only a desired state trajectory $\mathbf{X}_d \triangleq \mathbf{X} + \bar{\mathbf{X}}$:
\begin{subequations}
    \label{eq:nlp}
    \begin{align}
        \underset{\mathbf{X},\mathbf{U}}{\text{minimize }} & \mathbf{\bar{x}}^T_N\mathbf{Q}_N\mathbf{\bar{x}}_N + \sum_{i=0}^{N-1} (\mathbf{\bar{x}}^T_i\mathbf{Q}\mathbf{\bar{x}}_i + \mathbf{u}^T_i\mathbf{R}\mathbf{u}_i) \\
\text{subject to } \mathbf{x}_{i+1} & = \mathbf{x}_i + h f(\mathbf{x}_i, \mathbf{u}_i), \ \boldsymbol{\lambda}_{n,i} \geq 0, \ g(\mathbf{u}_i) \leq 0 \  \forall i \label{eq:nlp_friction}.
    \end{align}
\end{subequations}
where $\mathbf{Q}_N$, $\mathbf{Q}$, and $\mathbf{R} $ are weight matrices that adjust the penalization of the state deviation and the control effort. Either friction model can be incorporated by replacing $g(\mathbf{u}_i) \leq 0$ by the corresponding constraints, \eqref{eq:tangent_cc} or \eqref{eq:tangent_stc}. To exploit the structure of the problem, we use a large-scale sparse solver that allows warm-starting, i.e., SNOPT~\cite{snopt}. At each time step during execution, the initial control input $\mathbf{u}_0$ is executed, and the shifted solution is used as the initial guess for re-planning in a receding-horizon fashion.

 \begin{figure}[h]
  \centering
  \subfigure[The MIP-based control from \ref{sec:quasi_static_pushing}.]{
    \includegraphics[width=0.45\textwidth]{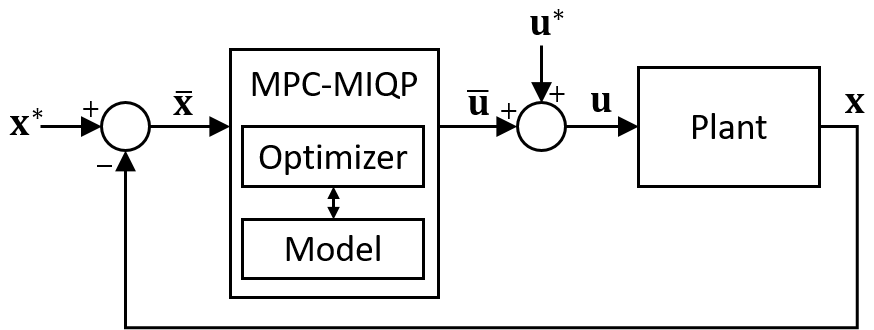}
    \label{fig:diag_miqp}
  }
  \hspace{.2cm}
  \subfigure[The NLP-based controller.]{
    \includegraphics[width=0.4\textwidth]{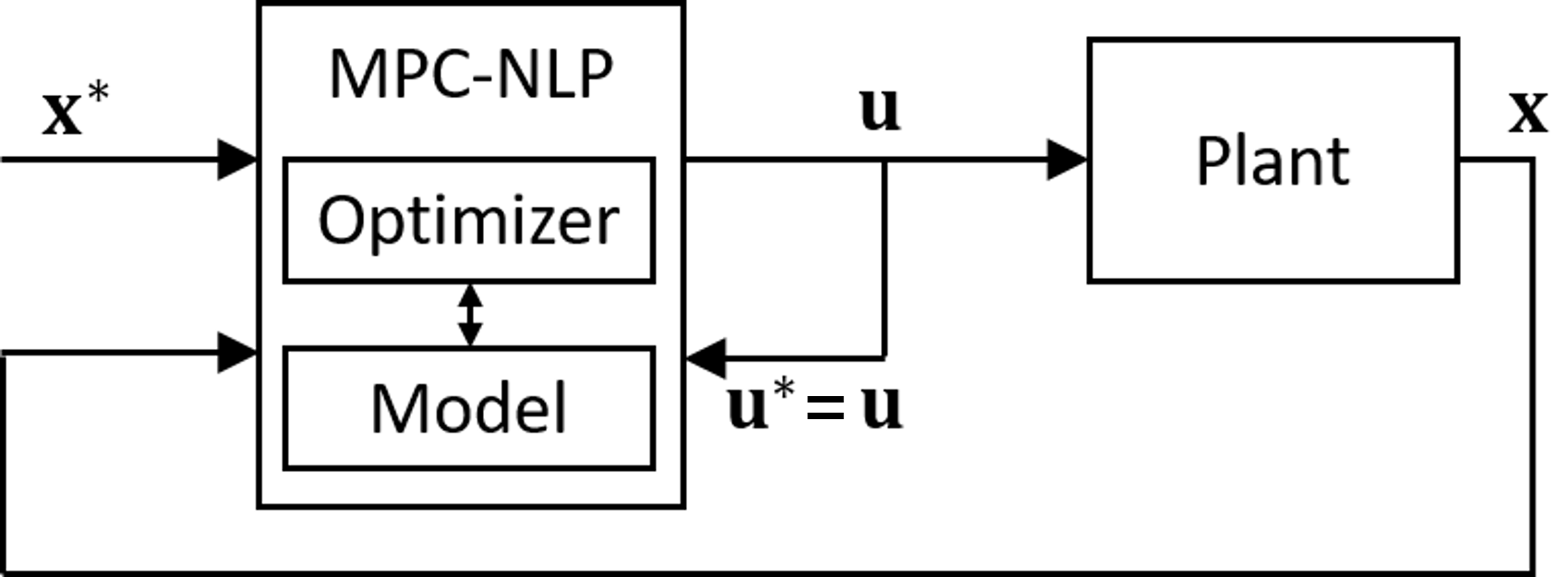}
    \label{fig:diag_nlp}
  }
  \caption{The block diagrams for the quasi-static pushing controllers.}
  \label{fig:block_diagrams}
\end{figure}

This formulation also differs from the MIP approach as the optimization variables are the absolute state and control values instead of their variations, as shown in Fig.~\ref{fig:block_diagrams}. A significant advantage of this approach is that a predefined contact mode schedule and control trajectory are unnecessary meaning the NLP-based approach can be used to plan a contact-interaction trajectory given only a desired state trajectory and using a trivial initial guess where the robot and the object are static.

%% file: sections/4_results.tex
\section{Experiments}
For the software implementation, the dynamics and the analytical gradients~\cite{carpentier2018analytical} are calculated using Pinocchio~\cite{pinocchio}. The NLP solvers IPOPT~\cite{ipopt} and SNOPT are interfaced using IFOPT~\cite{ifopt}. The MIP is solved using Gurobi~\cite{gurobi}. The computations were run on an AMD Ryzen 7 2700x processor.


\subsection{Dynamic Pushing}
Two non-prehensile manipulation tasks are designed to evaluate the proposed methodology: a translation-only task for the CA method and a rotation-only task for the friction models. We use a standard 6-DOF robot arm, UR3e from Universal Robots, and a cube with a side length of $0.1$~m and a mass of $0.6$~kg. The friction coefficient is 0.6 for contact pairs between the box and robot and 0.2 between the box and ground. The weights in \eqref{eq:optimization_problem} are set as $w_1 = 1$ and $w_2 = 10^4$. The size and number of time steps are selected as $h=0.05$~s and $N = 20$.

\subsubsection{Contact Activation}
Our proposed STC-based automatic CA is tested against using all contact constraints. In this task, the robot needs to translate the box into a target position. Four contact pairs are defined between each vertical surface of the box and the robot's end effector. The box is initialized at $(0.5, 0.0)$~m. 100 random target positions are sampled uniformly from the range of $([0.5, 1.0]$, $[-0.25, 0.25])$~m.

The left half of Table~\ref{table:pushing} shows the computation time statistics for converged runs and the success rates with and without CA for both solvers. When using IPOPT, significant advantages regarding computational time, on average a reduction of over $70\%$, and more than twice the success rate are achieved with CA. With SNOPT, the success rate is lower, however the overall performance with CA is still better. This is significant as it can enable automating contact candidate assignment from perception data. That is, all detected surfaces can be assigned as contact candidates since the number of contact pairs would not be an issue when the constraints are handled efficiently on the fly.

\begin{table}[]
\renewcommand{\arraystretch}{1.3}
\centering
\caption{Comparative results against baselines for 100 random simulation experiments.}
\label{table:pushing}
\begin{tabular}{|l|l|c|c||c|c||c|c||c|c|}
\hline
\multicolumn{2}{|l|}{Solver} & \multicolumn{2}{c||}{IPOPT} & \multicolumn{2}{c||}{SNOPT}  & \multicolumn{2}{c||}{IPOPT} & \multicolumn{2}{c|}{SNOPT} \\ 
\cline{1-10} 
\multicolumn{2}{|l|}{Condition}     & \multicolumn{1}{c|}{w/ CA} & \multicolumn{1}{c||}{w/o CA} & \multicolumn{1}{c|}{w/ CA} & \multicolumn{1}{c||}{w/o CA} & \multicolumn{1}{c|}{STCs} & \multicolumn{1}{c||}{CCs} & \multicolumn{1}{c|}{STCs}   & \multicolumn{1}{c|}{CCs} \\
\hline
\multirow{3}{*}{Time [s]} & Avg. & \textbf{5.19}  & 18.91 & \textbf{24.53} & 27.93
                                 & \textbf{20.09} & 25.85 & \textbf{6.45} & 18.05 \\
\cline{2-10}              & Min. & \textbf{1.39}  & 3.92  & \textbf{1.46}  & 1.92
                                 & 4.39          & \textbf{1.96} & \textbf{0.78} & 0.88 \\
\cline{2-10}              & Max. & \textbf{22.99} & 74.00 & \textbf{77.89} & 87.18
                                 & \textbf{50.67} & 94.70 & \textbf{49.71} & 85.52 \\
\hline
\multicolumn{2}{|l|}{Success Rate}  & \textbf{98\%} & 37\%  & \textbf{57\%} & 32\% 
                                    & \textbf{99\%} & 90\%  & \textbf{74\%} & 38\% \\
\hline
\end{tabular}
\end{table}
\vspace{-1cm}

\subsubsection{Friction Model}
We compare the STC- and CC-based friction models for the discovery of tangential forces considering a rotation-only task where a box that is attached to the ground via a revolute joint is commanded an orientation. A contact pair is defined between the robot's end effector and the center point of the box's surface that faces the robot. Hence, the robot can exert forces only on the center point of the vertical surface. Since this point aligns with the box's center of mass, frictional forces must be employed to complete the task. The box is initialized at 0~rad, and 100 goals are uniformly sampled from $[-1, 1]$~rad.



The right half of Table~\ref{table:pushing} shows the computation time statistics for converged runs and the success rates for both models and solvers. On average, the STC-based model outperforms the baseline both in speed and robustness. We believe that this is owing to the reduction of the number of binding constraints. In other words, the Coulomb friction law -- which, in fact, consists of \textit{if} conditions -- can be more naturally represented using STCs. Whereas, the CC-based model is trying to capture the same by using excessive bi-directional conditions.



\subsection{Quasi-Static Planar Pushing}
We compare the NLP- and MIP-based controllers through simulation experiments using the following parameters for both: $h = 0.3$~s, $N = 35$, $\mathbf{Q} = 10\ diag(3,3,0.1,0)$, $\mathbf{Q}_N = 2000\ diag(3,3,0.1,0)$, and $\mathbf{R} = 0.5\ diag(1,1,0.01)$. For the MIP-based controller, aggregated contact mode sections are used to accelerate the problem solving, as in \cite{hogan2020reactive}.\footnote{The number of steps associated with each section is $\{1, 5, 5, 5, 5, 5, 5, 4\}$ and the contact mode weight matrix is $\mathbf{W}=0.1\ diag(0, 1, 1, 1, 1, 1, 1, 1)$.}

\begin{figure}[tb]
  \centering
  \subfigure[Computational time comparison.]{
    \includegraphics[width=0.46\textwidth]{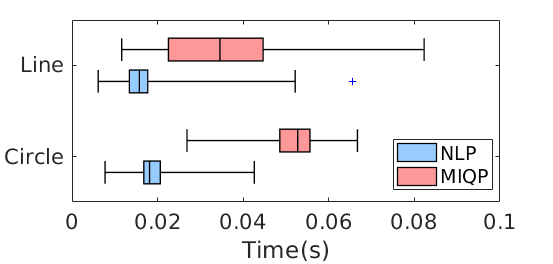}
    \label{fig:time_box_plot}
  }
  \subfigure[Tracking error comparison.]{
    \includegraphics[width=0.46\textwidth]{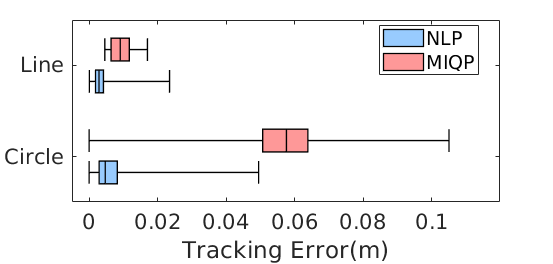}
    \label{fig:error_box_plot}
  }
  \caption{Performance for the quasi-static pushing simulations.}
  \label{fig:box_plot}
  \vspace{-0.5cm}
\end{figure}

\begin{figure}[tb]
  \centering
  \subfigure[]{
    \includegraphics[width=0.42\textwidth]{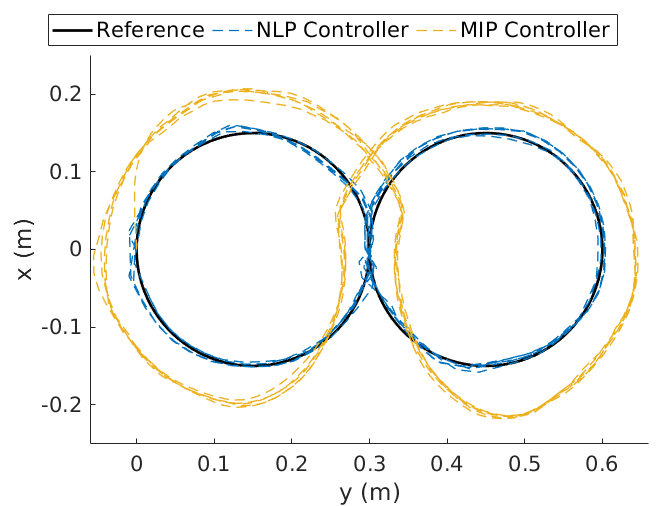}
    \label{fig:sim_circle}
  }
  \subfigure[]{
      \includegraphics[width=0.45\textwidth]{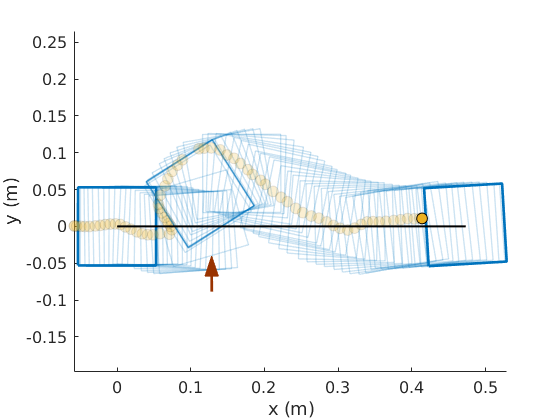}
      \label{fig:line}
  }
  \caption{(a) NLP- vs. MIP-based controllers in simulation for tracking an 8-shaped trajectory for 5 consecutive laps. (b) Hardware experiments for straight-line tracking with perturbation, i.e., the red arrow.}
  \vspace{-0.25cm}
\end{figure}

First, we consider a straight-line tracking scenario for a distance of $0.5$~m at a reference velocity of $0.05$~m/s. A feasible initial control trajectory is provided to both controllers. 100 simulations are performed with randomly-perturbed initial conditions that are drawn uniformly in the range $\pm [0.03, 0.03, 0.4, 0.025]$. The computation time and tracking accuracy results are shown in Fig.~\ref{fig:box_plot}. Overall, both controllers can handle the perturbations and accomplish the task. In this case, the NLP-based controller outperforms the MIP-based controller in all terms except for the maximum error bound. Second, an 8-shaped trajectory that consists of two circles of radius 0.15~m is tracked for 5 consecutive laps at a uniform desired angular velocity of 0.3~rad/s. However, in this case, a trivial infeasible control sequence is provided to the controller. The quantitative performances and the trajectories are shown in Figs. \ref{fig:box_plot} and \ref{fig:sim_circle}. The results show that the MIP-based controller cannot track the circular path accurately without a feasible nominal control sequence. This is likely caused by the linearized dynamics becoming ineffective for large deviations from the nominal trajectory. On the other hand, the NLP-based controller achieves precise tracking. Furthermore, the NLP can be solved within 0.02~s on average, which is promising for real-time applications.

\subsubsection{Hardware Experiments}
\begin{figure}[tb]
  \centering
  {\includegraphics[width=0.9\textwidth]{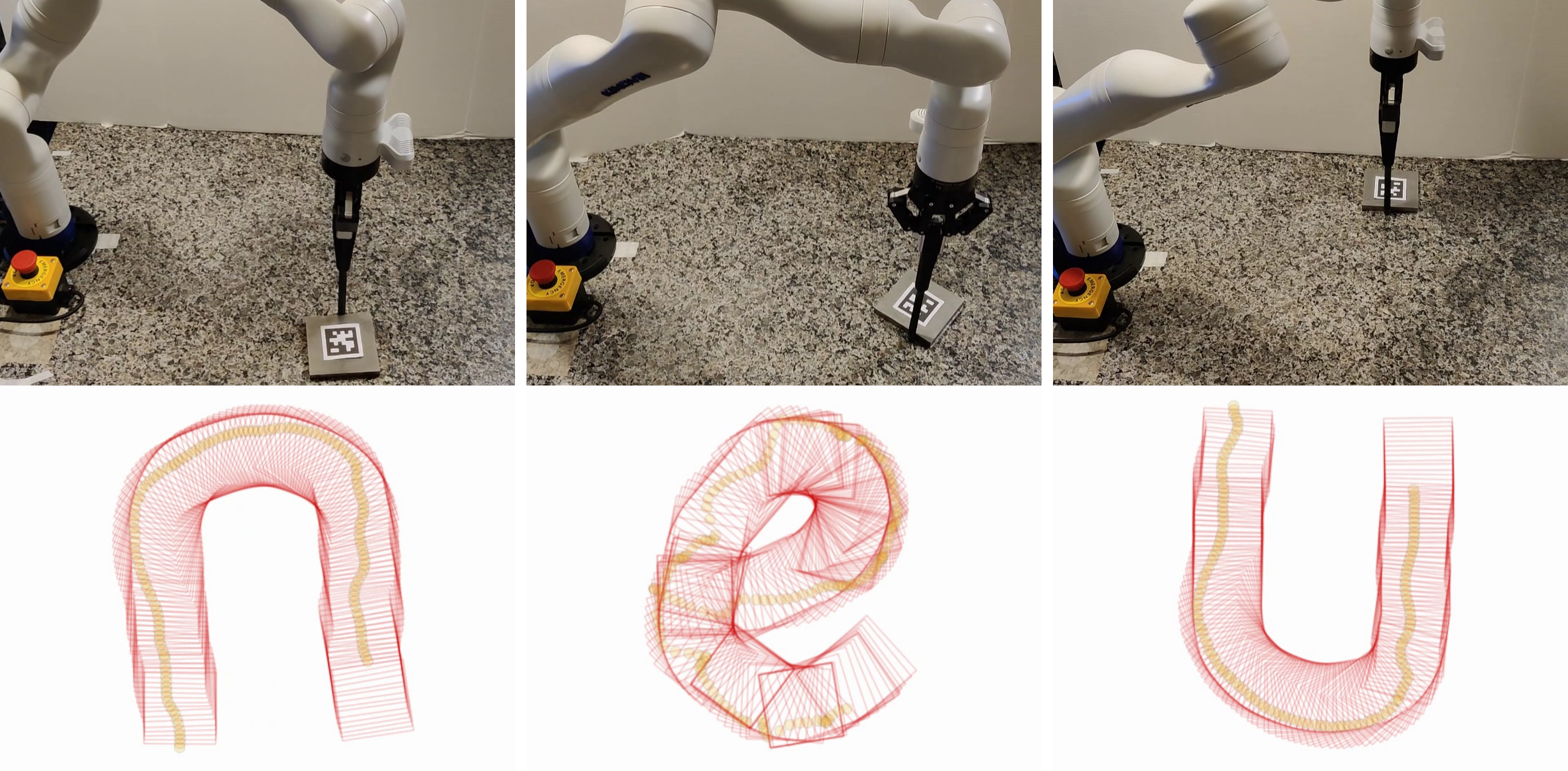}}
  \caption{Tracking of randomly-selected trajectories to demonstrate the generalization of the NLP-based controller. (Top) Snapshots from hardware experiments and (Bottom) the resulted trajectories for ``n"-, ``e"-, and ``u"-shaped trajectories.}
  \label{fig:neu}
\end{figure}

The proposed framework is also validated using the hardware setup shown in Fig.~\ref{fig:neu}. A Kinova Gen3 robot arm is used with a 3D printed pusher to manipulate a square metal block of length 0.1~m and mass 1~kg. The friction coefficient between the block and the ground is estimated as 0.4. As the robot arm is position controlled, the control inputs solved from the optimization are mapped to pushing velocity through a linear mapping~\cite{hogan2020reactive}. A camera mounted on the arm and AprilTag 3~\cite{kallwies2020determining} are used to estimate the slider's pose. Without occlusion, the localization error is within $0.3$~mm. The controller parameters used in simulation are maintained except that $\mathbf{R}$ is reduced to $0.1\ diag(1,1,0.01)$ to accommodate planning from scratch.

\begin{figure}[tb]
  \centering
  \subfigure[5 laps w/o external perturbation.]{
      \includegraphics[width=0.45\textwidth]{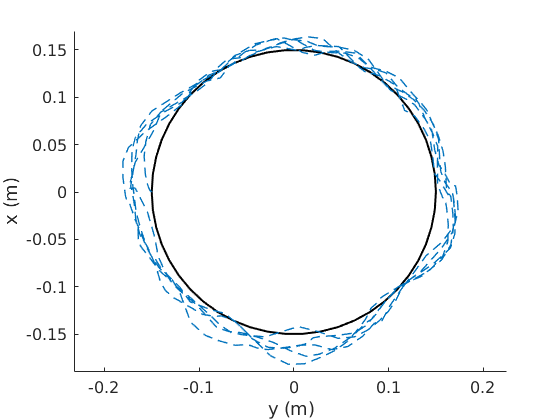}
      \label{fig:circle_wo_perturbation}}
  \subfigure[1 lap w/ external perturbation.]{
    \includegraphics[width=0.45\textwidth]{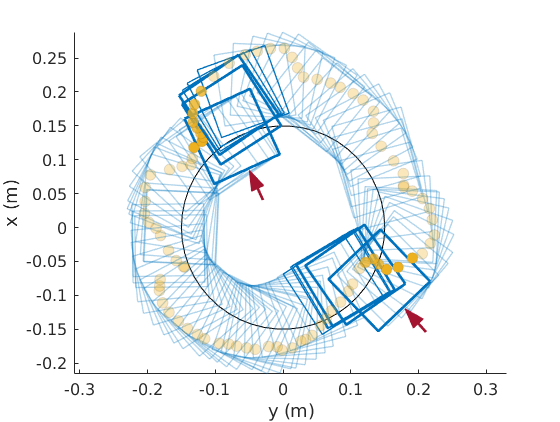}
    \label{fig:circle_w_perturbation}}
  \caption{Circle tracking on hardware where black and blue lines denote the reference and actual paths, yellow dots the pusher position, and red arrows the external perturbations.}
  \vspace{-0.25cm}
\end{figure}

We first test this framework for tracking a straight line and a circle under external forces. The result for tracking the straight line is shown in Fig.~\ref{fig:line}. It is seen that the controller reacts quickly and drives the slider back to the reference trajectory by adapting friction modes as necessary upon perturbation. Fig.~\ref{fig:circle_wo_perturbation} shows the result of tracking a circle of radius 0.15~m for 5 consecutive laps at a uniform desired reference angular velocity of 0.3~rad/s. In this case, an average tracking error of 0.019~m is obtained with an average computational time of 0.0255~s. While the controller is able to achieve stable circle tracking for 5 laps, the actual path constantly deviates from the reference path on the left and right end of the circle. We attribute this to the variations of task-space control performance and object tracking accuracy across the workspace. Referring to the hardware setup in Fig.~\ref{fig:neu}, the right end and left ends of the reference circle are the positions where the robot gets close to its operation range limits. The results for circle tracking with external perturbations are shown in Fig.~\ref{fig:circle_w_perturbation} proving that the controller can reject disturbances even when state deviations are large.

To further evaluate the proposed controller's generalization, we test it for randomly-selected complex trajectories. The trajectories of shapes ``n", ``e", and ``u" are used as references. Snapshots from the hardware experiments are depicted in the top row of Fig.~\ref{fig:neu}, and the bottom row shows the resulting trajectories.\footnote[1]{Please see the accompanying video for the full motion.} Doing this using the baseline would be challenging due to the high computational cost of MIP~\cite{hogan2016feedback} and the need for re-training for the learning-based model~\cite{hogan2020reactive}. We obtain average tracking errors of $0.0313$~m, $0.0332$~m, $0.0310$~m for the ``n", ``e", and ``u" shapes with an average computational time of $0.017$~s. It is noteworthy that despite the low-accuracy tracking, we have not observed any performance issues qualitatively; nevertheless, the quantitative tracking errors should be affected by this. We believe that a more accurate motion-capture system would improve the controller's performance.

%% file: sections/5_conclusion.tex
\section{Conclusion}
We have presented two key innovations for contact-implicit planning using STCs. First, we have introduced an automatic contact constraint activation method that enables the system to disregard unpromising contact pairs on the fly. We believe this will enable systems to explore a much larger solution space while maintaining efficiency and thus could prove important in applications such as manipulation in clutter. Second, we have proposed a re-formulation of Coulomb friction. Both approaches have been evaluated for dynamic pushing scenarios in simulation against CC-based baselines and found to provide significant improvements in computation time and success rate. Finally, by leveraging the proposed friction model, we formulate the quasi-static planar pushing problem as a direct NLP that can be solved fast enough to be run as an MPC. The proposed controller has been compared to a MIQP-based approach and shown to achieve higher control frequency and better tracking accuracy in simulation. Moreover, the proposed method has been demonstrated on hardware for reactive planning of complex trajectories in real-time without any heuristics.